\documentclass[11pt,a4paper]{article}
\usepackage[hyperref]{acl2021}
\usepackage{times}
\usepackage{latexsym}
\usepackage{comment}

\usepackage{microtype}

\usepackage{graphicx}

\aclfinalcopy 
\def\aclpaperid{6} 

\title{Dorabella Cipher as Musical Inspiration}

\author{Bradley Hauer, Colin Choi, Abram Hindle, Scott Smallwood, 
Grzegorz Kondrak \\
University of Alberta, Edmonton, Canada \\
{\tt {\{bmhauer,cechoi,hindle1,ssmallwo,gkondrak\}}@ualberta.ca}
}

\date{}

\begin{document}
\maketitle

\newcommand{\norvigc}{\textsc{HillClimbC}}
\newcommand{\pitchDuration}{pitch/duration}

\begin{abstract}
The Dorabella cipher
is an encrypted note written by English composer Edward Elgar, 
which has defied decipherment attempts for more than a century. 
While most proposed solutions are English texts, 
we investigate the hypothesis that Dorabella represents enciphered music. 
We weigh the evidence for and against the hypothesis,
devise a simplified music notation, 
and attempt to reconstruct a melody from the cipher. 
Our tools are n-gram models of music
which we validate on existing music corpora 
enciphered using monoalphabetic substitution. 
By applying our methods to Dorabella,
we produce a decipherment with musical qualities, 
which is then transformed via artful composition into a listenable melody.
Far from arguing that the end result represents the only true solution, 
we instead frame the process of decipherment 
as part of the composition process.
\end{abstract}

\section{Introduction}
\label{sec:introduction}


The Dorabella cipher 
(henceforth, simply {\em Dorabella})
is an encrypted note 
sent by Edward Elgar, the composer of the ``Enigma Variations'',
to his friend Dora Penny in 1897
\cite{santa2010}.
While many cryptography researchers have assumed that the 
underlying message is an English text,
it has also been hypothesized that it may encode music,
since Elgar was a composer and a music teacher.
This raises several interesting questions.
Is it possible to find evidence for or against the music hypothesis?
What kind of music notation could be devised with only two dozen
possible distinct symbols?
How would a musical decipherment compare to the proposed textual
decipherments?


In this paper,
we attempt to answer these questions 
in a principled manner, by 
using n-gram language models
derived from collections of transcribed music.
However, we also approach musical decipherment 
as a creative process.
We demonstrate this technique on Dorabella, 
producing a decipherment that has musical qualities, 
transformed via artful composition into a listenable melody. 
While prior work typically pursues a single correct decipherment, 
we instead adopt a creative approach
of converting ciphers into music,
which might lead to composition of new works.

This paper has the following structure:
In Section~\ref{background} we provide background on n-grams, perplexity and monoalphabetic substitution ciphers.
In Section~\ref{prior}, we discuss prior work. 
In Section~\ref{methods},
we describe our methodology, including datasets 
for training language models.
In Section~\ref{result},
we describe our results on encrypted melody samples. 
In Section~\ref{compose},
our highest-scoring decipherment of Dorabella as a melody
is used as inspiration to compose a new work.

\begin{figure}[t]
  \centering
  \includegraphics[width=0.8\columnwidth]{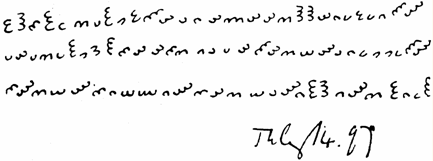}
  \caption{The Dorabella Cipher.}
  \label{fig:dorabella}
\end{figure}

\section{Background}
\begin{figure*}[th]
\begin{center}
\begin{scriptsize}
\begin{verbatim}
A2 E3 B2 A3 A1 C2 G1 A3 D1 H2 B3 F2 F1 B1 F2 C3 F2 F2 C2 E3 E3 F2 B1 H1 H2 H1 C1 B3 F3
G1 F2 G1 C2 H1 A3 D1 D2 A3 B2 F2 F2 B2 C2 C1 F1 G1 F2 B3 F2 C2 G2 F3 F1 B1 H1 D1 D1 H1 B3 F3
B2 F3 C2 G2 F3 B2 B1 G2 G3 C1 F3 B2 F2 C2 G2 F1 F3 C1 A3 E3 C1 F3 C2 A3 B1 H1 A3
\end{verbatim}
\end{scriptsize}
\caption{Our transcription of Dorabella 
which encodes the orientation and semicircle count of each symbol.}
\label{bumps}
\end{center}
\end{figure*}

\label{background}


Substitution ciphers, their properties, and cryptanalysis techniques
have been studied for centuries \cite{singh2011}.
A \emph{monoalphabetic substitution cipher} 
enciphers a \emph{plaintext} by applying a 1-to-1 mapping of symbols
to each character token, 
producing a \emph{ciphertext} which has a length 
equal to the length of the plaintext. 
The symbol mapping function is called the \emph{key}.
Given the key, reversing the encipherment process
and recovering the plaintext is trivial:
simply apply the inverse of the key to each ciphertext symbol.
The process of recovering the plaintext when the key is \emph{not} given
is called 
\emph{decipherment}.
Common measures of the decipherment success are: 
(1) {\em decipherment accuracy}, which is 
the percentage of correctly recovered symbols in the ciphertext;
and (2) {\em key accuracy}, which is 
the percentage of correctly mapped symbols in the cipher alphabet.
Decipherment accuracy is typically higher than key accuracy, 
because more frequent symbols are more likely to be deciphered correctly.

Computational decipherment methods 
are based on heuristic search algorithms
guided by statistical n-gram language models \cite{nuhn2013,hauer2014}.
An \emph{n-gram language model}
estimates the probability of a token in a sequence
based on the previous n$-1$ tokens.
If the token is near the beginning of the string,
a special start token is used in place of the missing prior tokens.
Through repeated applications of this model, 
the probability of the entire sequence can be estimated.
{\em Perplexity} is a function of probability,
which measures the ability of a statistical model to
predict a particular sequence.
Lower perplexity indicates that the model is 
``less surprised'' by the data, and so is said to be a better fit.
Tokens may be characters or words in natural language,
or symbols used in music notation.
For compatibility with prior work 
which models sequences of characters in natural language, 
we refer to a language model over music notation 
as a \emph{character language model}.

\section{Prior Work}
\label{prior}

This section describes prior works that use n-gram models for
composition, and prior attempts to solve the Dorabella cipher.

\subsection{N-Gram models for composition}

N-gram models have been applied to study the structure of
music~\cite{manzara1992entropy} and to compose music. N-gram models in
music research and composition range from serial notes, to chords, to
pitch and duration pairs~\cite{lo2006evolving,wolkowicz2008n}, and more
complicated structures~\cite{Mccormack96grammarbased}.

\newcite{manzara1992entropy} investigate the entropy of
music from an n-gram perspective.
They test how well people can
guess the next note, and compare that to n-gram models of Bach's
four part {\em Chorale}. 
They report that
people outperform n-gram models, 
and that both people and n-gram models have relatively consistent performance.

\newcite{Mccormack96grammarbased} employs n-grams and similar
Bayesian structures to compose music. His focus was on Markov chains,
which are related to n-gram language models and perplexity estimations.

\newcite{lo2006evolving} combine genetic algorithms and n-gram
language models to evolve musical sequences. The n-gram models
act as fitness functions to guide the creation of musical sequences
that have lower perplexity given an n-gram language model. 
In addition, they 
use their models to identify composers.

\newcite{wolkowicz2008n} also use n-grams 
to identify composers. 
They process MIDI files and produce n-grams of pitch and duration tuples. 
They achieve up to 84\% accuracy at identifying
composers using a large corpus of 10000 notes of each composer's work,
and about 54\% accuracy when using only 100 notes,
which is at a similar level of accuracy as \newcite{lo2006evolving}.

\subsection{Dorabella Cipher}

The earliest computational attempt 
at solving the Dorabella Cipher that we are aware of
is that of \newcite{Sams}.
He applies statistical analysis
based on character frequencies
and brute force cryptanalysis.
The work considers the assumptions that the cipher
encrypts English text which may be partly phoneticized,
is not strictly monoalphabetic,
and may involve multiple layers of encryption.
The author ultimately proposes the following solution to the cipher:
\emph{
``Larks! It's chaotic, but a cloak obscures my new letters, a, b.
I own the dark makes E. E. sigh when you are too long gone.''
}

\newcite{santa2010} provide an overview 
of Elgar's work on cryptography,
focusing on the ``enigma'' that he implied to be hidden
within his musical piece \emph{Variations on a Theme}.
They note the connections Elgar made between that piece and Dorabella,
neither of which has been conclusively solved.
despite this and other ``hints'' from Elgar. 
In particular, they raise the possibility 
of mathematical concepts being used in Dorabella,
specifically the constant $\pi$,
as well as the encoding of scale-degrees with numerical values.

As well-known techniques, such as frequency analysis,
have not proven effective on Dorabella,
\newcite{schmeh} 
proposes to consider less common techniques.
These include vowel detection and 
a frequency-based consonant identification method.
The author applies these techniques both to Dorabella,
and on a control plaintext.
He does not propose a solution to Dorabella,
but demonstrates that these methods
can distinguish between vowels and consonants in the control cipher.
With the same techniques, 
he attempts to
identify some Dorabella symbols as vowels or consonants.
He also notes that certain statistical properties of Dorabella
are consistent with English text.

\newcite{packwood2020} proposes a natural language solution 
to Dorabella.
The method is complex, and involves breaking the cipher
into discrete blocks, among which patterns can be observed,
and an elaborate system of transposition.
The author further speculates
that the cipher also conceals a musical composition,
but makes no attempt at 
a musical decipherment.

\newcite{hauer2021} experiment with several monoalphabetic
substitution cipher solvers to decipher music.
They rely on a corpus of Bach and Elgar MIDI files, and try to decipher
synthetic music ciphers 
using a {\pitchDuration} language model,
but the results are quite low compared to textual ciphers.
They conclude that it is unlikely that Dorabella
represents music encoded using an alphabet of pitch and duration.
%



\section{Methodology}
\label{methods}

In this section,
we describe our methodology, including datasets for training language models.

\subsection{Transcribing Dorabella}



The first step is to render Dorabella into a machine-readable form.
In order to establish such a transcription,
we compared five different manual transcriptions attempts,
including 
\newcite{schmeh},
\newcite{benzedrine},
\newcite{massey},
as well as transcriptions by two of the authors of this paper.
A majority consensus transcription 
is shown in Figure~\ref{bumps}. 
It consists of 87 tokens made of an uppercase letter
encoding the symbol orientation,
followed by the number of semicircles.
There are 8 possible orientations (A-H), 
while the number of semicircles ranges from 1 to 3.

\subsection{Pitch-Duration Dataset and Encoding}
\label{sec:pitch}




We use the music dataset created by \newcite{hauer2021}.
The dataset was created from MIDI files,
a form of digitally representing musical composition
which encodes pitch, pitch amplitude, and duration over a timeline, 
usually including metric and tempo information.
The files represent music from both Elgar and Bach.
The Elgar data consists of 29 files containing a total of 1.2M notes,
while the Bach data consists of 295 files containing 3.7M notes.
We include the Bach data due to the relatively small size of the Elgar corpus;
this 
increases the total size of our data by a factor of four.
Each dataset is divided into training and testing splits.
This is done to ensure that experimental results are generalizable
to data not used to provide statistical information for the
models used by the decipherment algorithms.
The test set is further divided into 87-note sequences,
the same length as Dorabella.

\newcite{hauer2021} assume that enciphered music must, 
before encipherment,
be represented in some serial, symbolic notation.
To this end, they transpose all music into the key of C major,
and use only one octave.
All symbols except notes (e.g. rests) are removed.
All notes are normalized to one of three durations:
quarter note, shorter than a quarter note, or longer than a quarter note.
Further, all notes were normalized to one of the eight most frequent notes:
A, B, C, D, E, F, F$\sharp$, and G.
Thus, just as each Dorabella cipher symbol
has one of three semicircle counts and one of eight orientations,
giving a theoretical vocabulary of 24 symbols,
the encoding assigns to each symbol
one of three durations and one of eight notes,
yielding 24 distinct symbols.
While there is much more information encoded in musical notation,
we are constrained by the 24-symbol alphabet of the cipher.
For example,
if we assumed that some cipher symbols represent rests,
we would need to further reduce the already limited
range of notes that the cipher can represent.
%
While this encoding was designed to match the form of the
Dorabella cipher,
we present a more principled approach in Section~\ref{dataset_melody}.

\subsection{English Dataset}


To assess the ability of our statistical models to fit music,
we induce models of both music and English,
and compare the fitness of our modelling method on different types of data.
%
We use the English language dataset of \newcite{hauer2021}, 
which is 
a subset of 
{\rm the letters of Jane Austen}. 
This corpus was deemed appropriate since 
it consists of written epistolary correspondences,
which is the hypothesized domain of Dorabella.
The text was first processed to remove all non-alphabetic characters,
including white space.
300 excerpts from the corpus were selected at random,
each consisting of a sequence of 87 characters. 
We use this set of 300 texts
for the perplexity measurement experiment 
described in Section~\ref{result}.

\subsection{Melody Dataset}
\label{dataset_melody}

We experiment with 
the CANTUS
corpus of folk music \cite{lacoste2012cantus}.
We conjecture that 
Elgar would be more likely to create something 
that reflected contemporary styles of folk music 
rather than the more complex, 
and often more chromatically adventurous music of his own.
This leads us to consider musical databases 
that are limited to a single line of music, 
as well as to simplify the issues 
around key signature, meter, rhythm, and other factors.
Our melody corpus 
reduces all examples to the common key of C. 
We also chose to not attempt to model 
rhythms, dynamics, articulations, and other components, 
looking mainly at pitch, and assuming 4/4 meter.  
%
There are other composition decisions which could have been made.
Given the extremely small set of symbols, and the short length of the cipher,
we necessarily had to make some simplifying assumptions,
and we did so based on
our intuitions regarding what setting would produce 
the most natural-sounding composition.

Our melody dataset is from the CANTUS Database of chants and melodies 
\cite{lacoste2012cantus,helsen2011report}, an online searchable database 
that encodes melodies as sequences of pitches
without including their durations.\footnote{An example 
melody \url{http://cantusindex.org/melody/msch001}}
Table \ref{tab:melody} shows the sources of melodies in the
CANTUS dataset and their average length. 
The melodies in CANTUS are monophonic, 
and most include notes in the range of F3 to D6. 
Since there are only 17 distinct notes in our subset of CANTUS,
we interpolate
the range from A3 to E6 to yield 24 symbols used to decipher Dorabella:
A3, B$\flat$3, B3, F3, G3,  
A4, B$\flat$4, B4, C4, D4, E4, F4, G4, 
A5, B$\flat$5, B5, C5, D5, E5, F5, G5, 
C6, D6, E6.
Because of the smaller alphabet and vocabulary of the melody dataset,
we expect it to have lower perplexity,
which should lead to better results than
with the dataset described in Section~\ref{sec:pitch}.

%


Our training corpus is created by randomly sampling
$467$ melodies without replacement.
Our datasets, code, and compositions are released at:
\url{https://zenodo.org/record/4764819}


\begin{table}[t]
\centering
\begin{tabular}{ccrr}
Name & Dataset & Melodies & Length\\
\hline
Gloria & mbos & 102 &  8.9 \\
Kyrie & mmel & 226 &  8.7 \\
Agnus Dei & mscb & 267 &  8.9 \\
Alleluia & msch & 409 & 34.9 \\
Hymn     & msta & 344 & 49.4 \\
Sanctus & mtha & 228 &  9.0 \\
\hline
     & All     & 1576 & \\
\end{tabular}
\caption{Melody datasets extracted from
CANTUS~\cite{lacoste2012cantus,helsen2011report}}
\label{tab:melody}
\end{table}

\subsection{Decipherment}


As our decipherment method for enciphered music,
we use the solver of \newcite{norvig-solver},
which we refer to as \norvigc{}.\footnote{\url{http://norvig.com/ngrams}}
We selected it for its effectiveness on deciphering 
monoalphabetic substitution ciphers,
even when word boundaries are not preserved in the cipher.
This is important, as our encoding of music has no analogy to word boundaries,
and no such boundaries are indicated in Dorabella.
The solver 
maximizes the probability of the decipherment
as estimated by a trigram character language model.
%
Starting from a random initial key,
\norvigc{} applies a hill climbing algorithm
as a heuristic search strategy. 
At each step, many successor keys are generated by applying permutations
to the current key; 
whichever successor gives the greatest increase in probability
(equivalently, the greatest decrease in perplexity)
becomes the key in the next iteration.
We run the algorithm for 4000 iterations, with 90 random restarts.
The decipherment with the lowest perplexity across all iterations is returned.

\section{Decipherment Results}
\label{result}


Table \ref{tab:elgarbachmelody} shows the decipherment results
on a test set of $300$ distinct melody samples,
sampled without replacement from the corresponding training set. 
Clearly, the results on the melody dataset are much better than those
on the {\pitchDuration} datasets,
which in turn are better for Bach than for Elgar.
The mean key accuracy across all examples in the melody dataset is 50\%, 
that is, half of the key symbols are correct. 
Approximately half
of ciphers were deciphered with 70\% decipherment accuracy or higher, 
and nearly one third of ciphers were deciphered entirely correctly.
This suggests that our approach is effective for melody decipherment.

\begin{table}[t]
  \begin{center}
    \begin{tabular}{l|c|c}
     \textbf{Source} & \textbf{Key Acc} & \textbf{Dec Acc} \\
      \hline
      {\pitchDuration} (Elgar) &  7.0\% & 12.0\%  \\
      {\pitchDuration} (Bach) & 26.5\% & 32.0\%   \\
      melody (CANTUS) & 50.0\% & 54.5\% \\
    \end{tabular}
  \end{center}
    \caption{\norvigc{} results on music ciphers of length 20,000.}
    \label{tab:elgarbachmelody}
\end{table}






One reason for
the lower accuracy on the {\pitchDuration} datasets
may be their quality.
The original MIDI files were created by multiple authors, 
leading to a low consistency in musical transcription.
In addition, 
the files are polyphonic; 
even for piano music, they often have separate channels for each hand. 
On the other hand,
our melody dataset 
is monophonic and consistently transcribed.


\begin{table}[ht]
\begin{center}
\begin{tabular}{l|c}
  Dataset & Average Perplexity \\
  \hline
English (Austen)          & 16.2 \\
\hline
{\pitchDuration} (Elgar) & 24.4 \\
{\pitchDuration} (Bach)  & 24.5 \\
melody (CANTUS)                & 5.6 \\ 
\end{tabular}
\end{center}
\caption{Average perplexity 
using a trigram character language model.
}
\label{tab:perp}
\end{table}

Another possible explanation for the divergent performance 
could be the encoding.
Table \ref{tab:perp} shows the perplexity of different datasets.
We used trigram language character models 
with modified Kneser-Ney smoothing and discounts. 
The relatively high perplexity values 
of the {\pitchDuration} datasets
suggests that the pitch-only encoding may be better
suited to modelling music
than the {\pitchDuration} encoding.
%
Indeed, based on these results, 
predicting the next note of a melody is easier than
than predicting the next English character in a sentence. 

\section{Composition from Decipherment}
\label{compose}

In this section we apply our algorithm to the Dorabella cipher,
and take the resulting melody as a basis for a composition.
In particular, we
manually analyze the output for musical content,
and modify it according to subjective musical tastes.
This creative process is guided by the familiarity with the composer's style,
and is not itself replicable.

\begin{figure}[htb]
  \centering
  \includegraphics[width=0.8\columnwidth]{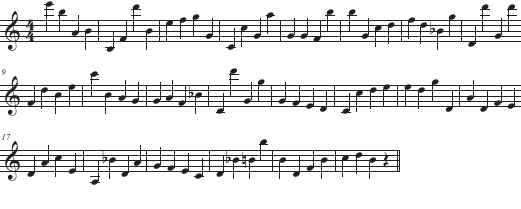}
  \caption{Note output from Dorabella}
  \label{noteD}
\end{figure}

Figure~\ref{melody_dec}
shows our highest-scoring decipherment of Dorabella assuming 4/4 time. 
Figure~\ref{noteD} depicts its musical transcription, which
was obtained by applying \norvigc{} 
with a language model derived from the melody dataset (467 samples).
%
This decipherment attempt has some interesting musical features.
The notes in Figure~\ref{noteD}
seem at times to imply logical harmonic progressions.
In the second half, there are even moments of motivic repetition, 
albeit not exact, which evoke a musical composition. 

%


\begin{figure}[t]
\begin{tiny}
E6 B5 A4 B4 B3 F4 D6 B4 E5 F5 G5 G4 C4 C5 G4 A5 G4 G4 F4 B5 B5 G4 C5 D5 F5 D5 B$\flat$4 G5 D4
D6 G4 D6 F4 D5 B4 E5 C6 B4 A4 G4 G4 A4 F4 B$\flat$4 C4 D6 G4 G5 G4 F4 E4 D4 C4 C5 D5 E5 E5 D5 G5 D4
A4 D4 F4 E4 D4 A4 C5 E4 A3 B$\flat$4 D4 A4 G4 F4 E4 C4 D4 B$\flat$4 B4 B5 B$\flat$4 D4 F4 B4 C5 D5 B4
\end{tiny}
\caption{A decipherment of Dorabella as melody.}
\label{melody_dec}
\end{figure}




\begin{figure}[tb]
  \centering
  \includegraphics[width=0.7\columnwidth]{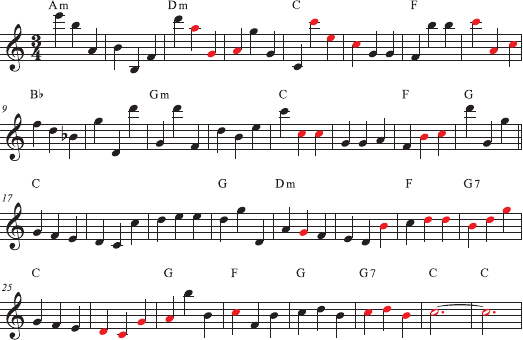}
  \caption{Adjusted output, with chords. 
  Notes which have been modified from the output 
  in Figure \ref{noteD} are color-coded red.}
  \label{noteC}
\end{figure}

After manual analysis 
we decided to realign the notes from the 4/4 meter to the 3/4 meter,
as this appears to fit better
the contours and implied harmonic progression.
Surprisingly, 
the melody seems to be match two 16-bar 3/4 phrases, 
except for a premature end in the second phrase.
Considering that we only use quarter-note rhythms, 
this could be an illusion,
but the resulting musical piece is intriguing.
In the spirit of the creative process,
we also decided to relax 
the strict matching of the decipherment symbols into notes.

Figure \ref{noteC} shows the final version of the output 
in which some notes (shown in red) have been altered or added in order
to create a cadential conclusion.
Interestingly, our altered 32-bar segment 
features a highly disjunct first 16-bar phrase, 
with an altered pitch (B-flat) implying a potential transposition, 
and a second phrase that is much more lyrical 
and based on smaller step-wise intervals, 
complete with the ``repeated motive.''
Upon adding the implied harmonic accompaniment, 
we can see that in some cases there even seem to be an implied V-I cadences,
such as between mm. 16-17, mm. 24-25, 
and the final added measures. 
Adding some phrasing and interesting timbres, 
as well as chords based on the implied harmony, 
gives us the audio rendering\footnote{\url{https://zenodo.org/record/4764819/files/fig3.wav?download=1}}
shown in Figure \ref{noteC}.

It is important to point out several caveats 
to this seemingly encouraging result.
%
First and foremost,
any analysis of musical composition necessarily has subjective elements.
Second,
we assume that rhythmic values are not encoded in the cipher,
and limit the decipherment to quarter-notes.  
It is also possible that 
the score may not be connected to common practice notation 
or even diatonic pitches at all.  
For example, these could be
referring to a very specific set of church bells,
or perhaps some other kind of instrument or sonic contraption,
or even just rhythm.

\section{Conclusions}

Although we do not claim to have solved the mystery of Dorabella, 
our process produced a listenable melody, 
which opens up interesting avenues of investigation.
%
In the future, we plan to experiment with 
different corpora and musical attributes, such as rhythm only.
Our approach
represents a creative way to generate new forms of musical melodies.
What seems certain is that Elgar's intention to confound
left us with a tantalizing riddle 
that invites further speculation in the future.


\nocite{Sams}
\nocite{packwood2020}

\bibliographystyle{acl_natbib}
\bibliography{dorabella}

\end{document}